\documentclass[11pt]{article}

\usepackage[preprint]{acl}

\usepackage{times}
\usepackage{latexsym}

\usepackage[T1]{fontenc}

\usepackage[utf8]{inputenc}

\usepackage{microtype}

\usepackage{inconsolata}

\usepackage{graphicx}
\usepackage{multirow}
\usepackage{colortbl}
\usepackage{tabularx}
\usepackage{placeins}
\usepackage{booktabs}
\usepackage{amsmath}

\usepackage{cleveref}
\usepackage[most]{tcolorbox}

\usepackage{xurl}
\hypersetup{breaklinks=true}

\newtcolorbox{systempromptbox}[1][System Prompt]{
  colback=blue!5,
  colframe=blue!60!black,
  title={#1},
  fonttitle=\bfseries,
  arc=2mm,
  boxrule=0.8pt,
  left=2mm,
  right=2mm,
  top=1mm,
  bottom=1mm,
  breakable
}

\title{Thinking with Visual Grounding}

\author{
Junkai Zhang \qquad
Yihe Deng \qquad
Kai-Wei Chang \qquad
Wei Wang \\
University of California, Los Angeles
}

\begin{document}
\maketitle
\begin{abstract}
Visual thinking should not only sound right; it should show its evidence. While recent vision-language models (VLMs) can produce natural-language reasoning traces, these traces often leave the supporting image regions implicit, making them hard to verify and difficult to supervise.
We introduce \emph{visually grounded thinking}, a reasoning process in which models interleave natural-language thoughts with explicit point or box groundings of the visual evidence used at each step. 
This lets the model express intermediate reasoning in language while grounding key objects in the image regions they refer to.
To train this behavior, we construct a scalable synthesis pipeline that distills correct visual reasoning traces, extracts the visual objects required by the traces, grounds them with a SAM3-based agent, and derives aligned point and box supervision from the resulting masks.
We further propose grounding-aware reinforcement learning, which combines answer correctness rewards with dense grounding rewards that score whether generated object references match the correct image evidence.
Across two counting benchmarks and four spatial reasoning benchmarks, adding visually grounded thinking to Gemma3-4B-IT consistently improves performance over the original model and the non-grounded thinking baseline.
On spatial reasoning, the visually grounded thinking 4B models match, and in some cases surpass, Gemma3-27B-IT from the same model family.
Our analysis shows that point grounding is well suited to counting, while box grounding benefits most from explicit grounding rewards on spatial tasks.
Overall, our results show that VLMs think better when their intermediate thoughts are tied to the image regions that make them true.\footnote{We release the \href{https://huggingface.co/datasets/JunkaiZ/TVG}{data}
and \href{https://github.com/Jun-Kai-Zhang/visually_grounded_thinking}{code}.}
\end{abstract}

\section{Introduction}

Language models have made strong progress on complex problem solving by producing explicit natural-language reasoning traces. In particular, R1-style reinforcement learning has shown that models can improve their ability to solve math, coding, and general reasoning problems through long textual thinking~\citep{Guo_2025}. This success has motivated analogous reasoning methods for vision-language models (VLMs): given an image and a question, the model can think in text before giving the final answer. Such a strategy has been shown to be effective for visual question answering~\citep{deng2025openvlthinkercomplexvisionlanguagereasoning,hu2026openvlthinkerv2}.



However, visual thinking differs from purely textual thinking because the evidence needed to solve a visual question is located in the image~\citep{zhu2016visual7w} and cannot be fully expressed in words. When humans answer visual questions, we often link our thoughts to concrete image regions, such as the person on the left, the cup near the table edge, or the object being counted~\citep{das2016humanattentionvisualquestion}. These visual references guide where attention should be directed and what task-specific information should be extracted~\citep{HAYHOE2005188}. In contrast, a pure natural-language reasoning trace may state that ``the red car is near the entrance'' or that ``there are three people holding umbrellas,'' but it does not identify which image regions support these claims. This makes the thinking hard to verify and supervise: a final answer may be correct even without the image, while the reasoning trace can still appear coherent and image-based~\citep{asadi2026mirage}. Thus, visual thinking requires not only step-by-step reasoning, but also explicit links between important reasoning steps and the correct visual evidence.

\begin{figure}[h]
  \centering
  \includegraphics[width=0.8\linewidth]{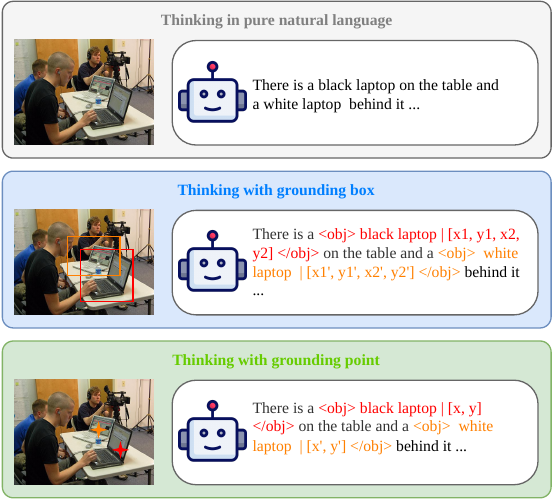}
  \caption{Thinking in pure natural language vs. visually grounded thinking in box mode and point mode.}
  \label{fig:illustration}
\end{figure}

We propose \emph{visually grounded thinking} to address this issue, as shown in \Cref{fig:illustration}. In this format, the model interleaves natural-language reasoning with explicit visual grounding. Whenever a reasoning step refers to an important visual object, the model outputs a coordinate-based grounding tag, using either a bounding box or a point, to identify the referenced object in the image. Natural language and spatial coordinates are combined in the thinking process: language describes the thoughts, while the coordinates specify the visual evidence that supports each step.

\begin{figure*}[!h]
  \centering
  \includegraphics[width=0.96\textwidth]{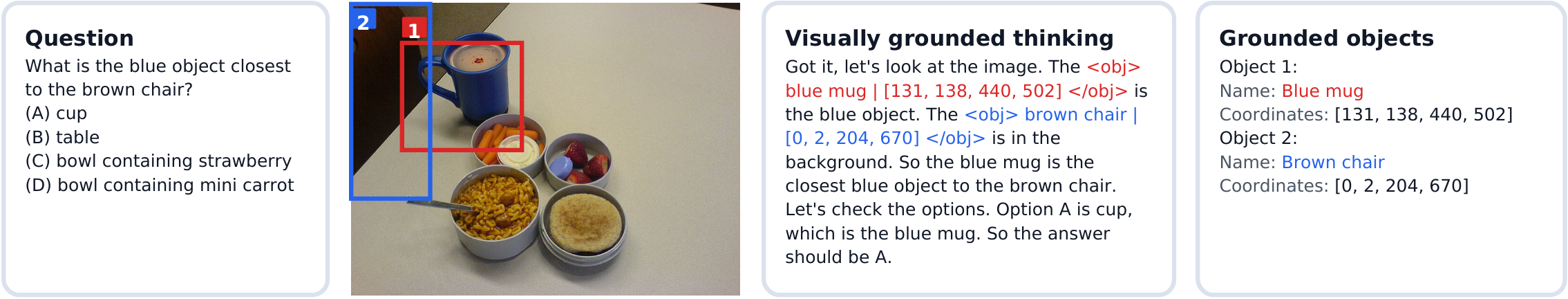}
  \caption{A real example of a visually grounded thinking model’s output in the evaluation benchmark.}
  \label{fig:eval-example}
\end{figure*}


Visually grounded thinking requires training data that supports both SFT and RL. Coordinate-annotated reasoning traces can teach the model to interleave language and grounding during SFT, but RL needs supervision at the level of each visual reference, since a rollout may rename objects, reorder reasoning steps, skip supervised entities, or ground additional useful evidence. We therefore build an automatic data synthesis pipeline around a SAM3-based grounding agent~\citep{carion2026sam3segmentconcepts}. Starting from visual questions, the pipeline obtains correct reasoning traces, extracts the visual objects used in the reasoning, represents each \emph{object} with a \emph{name} and a disambiguating scene \emph{context}, and grounds it with a run-length encoding (RLE) mask. These masks are used to construct both point and box-mode grounded reasoning traces, while the corresponding grounded objects are kept as structured supervision for grounding-aware RL.

We train models to perform visually grounded thinking with the synthetic data. The models are first cold-started with synthesized visually grounded reasoning traces, where important objects are grounded with either points or boxes. We then apply RL with the reward explicitly supervising the grounding quality. This stage jointly encourages answer correctness and precise grounding of the visual objects that support the model’s intermediate reasoning. An example output from a visually grounded thinking model is shown in \Cref{fig:eval-example}.


Our controlled experiments show that visually grounded thinking substantially improves counting and spatial relationship understanding. On spatial relationship benchmarks, our 4B visually grounded thinking models reach performance comparable to, and in some cases higher than, the 27B model from the same family. The grounding reward brings clear gains for box-mode grounded thinking, especially on spatial tasks where object extent and relative geometry are important. We also find that the two grounding interfaces have different strengths: point grounding performs better on counting, where instance-level localization is often sufficient, while point and box grounding are broadly comparable on spatial reasoning.

Our contributions are as follows:
\begin{enumerate}
\item We build a scalable data pipeline to synthesize visually grounded thinking data for both SFT and RL, centered on a SAM3-based agentic grounding system that extracts high-fidelity object masks as visual supervision.
\item We design a grounding reward that directly supervises whether the model grounds its intermediate visual references in the correct image evidence, supporting both box-mode and point-mode grounded thinking.
\item Through controlled experiments, we show that visually grounded thinking substantially improves counting and spatial reasoning. The grounding reward brings clear gains for box grounding, especially on spatial tasks. In addition, point grounding is particularly effective for counting and remains competitive with box grounding on spatial reasoning.
\end{enumerate}

\section{Related Work}

Early work on visually grounded thinking mainly uses grounding to locate the image regions needed for answering a question. Visual CoT~\citep{shao2024visual} introduces intermediate bounding boxes that highlight key regions, while UV-CoT~\citep{zhao2025unsupervised} reduces the need for human box annotations by learning from preferences over model-generated regions. Later work more tightly couples grounding with the reasoning trace. GCoT~\citep{wu2025grounded}, \citet{xia2025bootstrapping}, and Argus~\citep{man2025argus} generate grounding coordinates as step-level visual evidence, aiming to make the reasoning more faithful to the image and easier to check. More recent work further treats grounding as an active behavior: GRIT~\citep{fan2026grit} and ViGoRL~\citep{sarch2026grounded} train models to interleave natural language with visual coordinates through RL, and VGR~\citep{wang2025vgr} uses predicted regions for visual replay during inference. Our work follows this shift from region-of-interest selection to visually grounded thinking, and extends it with an explicit grounding reward that directly scores the visual grounding produced during thinking.

\section{Data Synthesis Pipeline}\label{sec:data}
\begin{figure*}[h]
\centering
\includegraphics[width=0.9\textwidth]{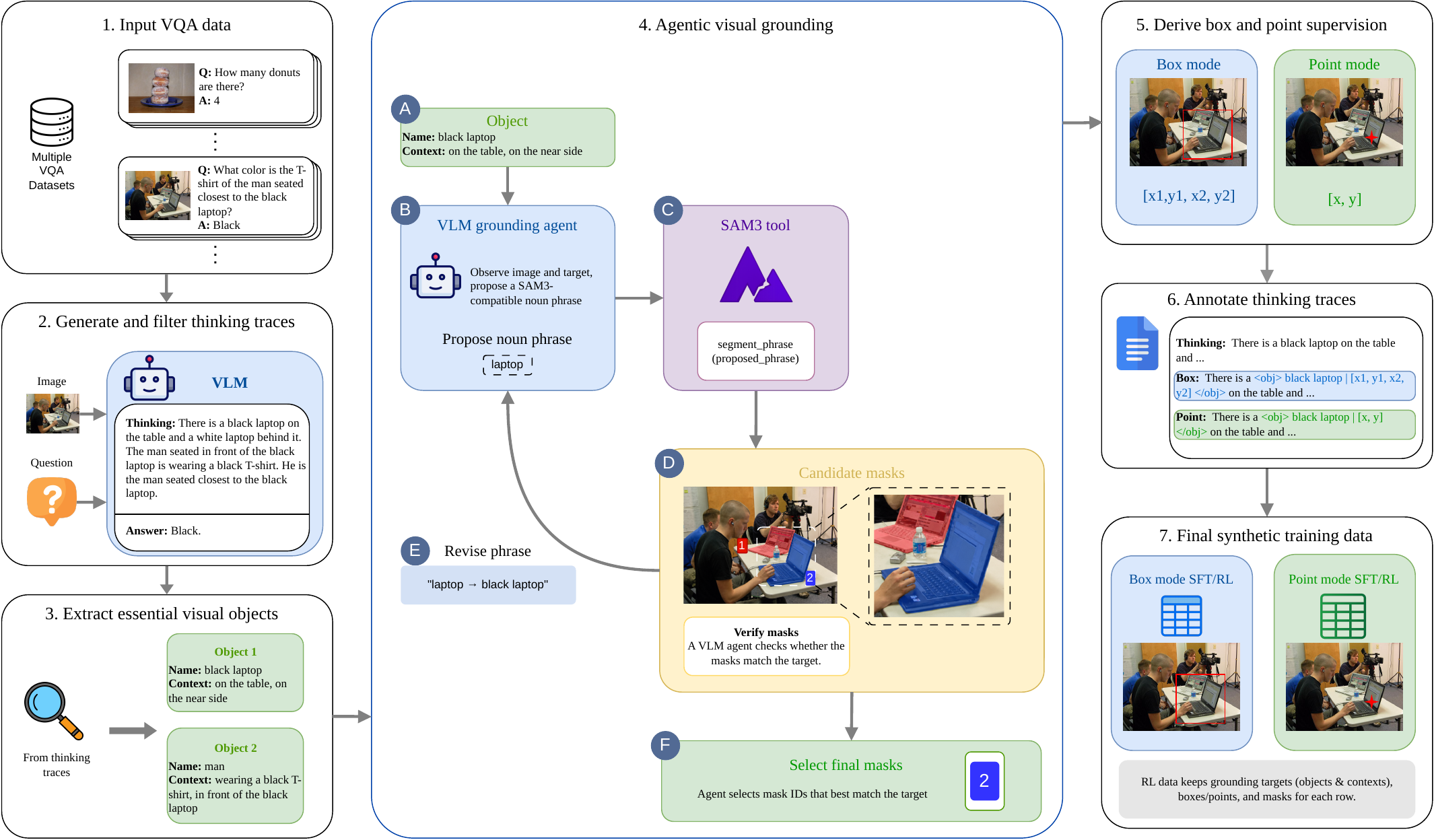}
\caption{Data synthesis pipeline. We distill reasoning traces, extract groundable visual evidence, ground those objects with an iterative SAM3-based agent, and write aligned box-mode and point-mode SFT and RL training data.}
\label{fig:data-synthesis-pipeline}
\end{figure*}
\begin{figure*}[h]
  \centering
  \includegraphics[width=0.9\textwidth]{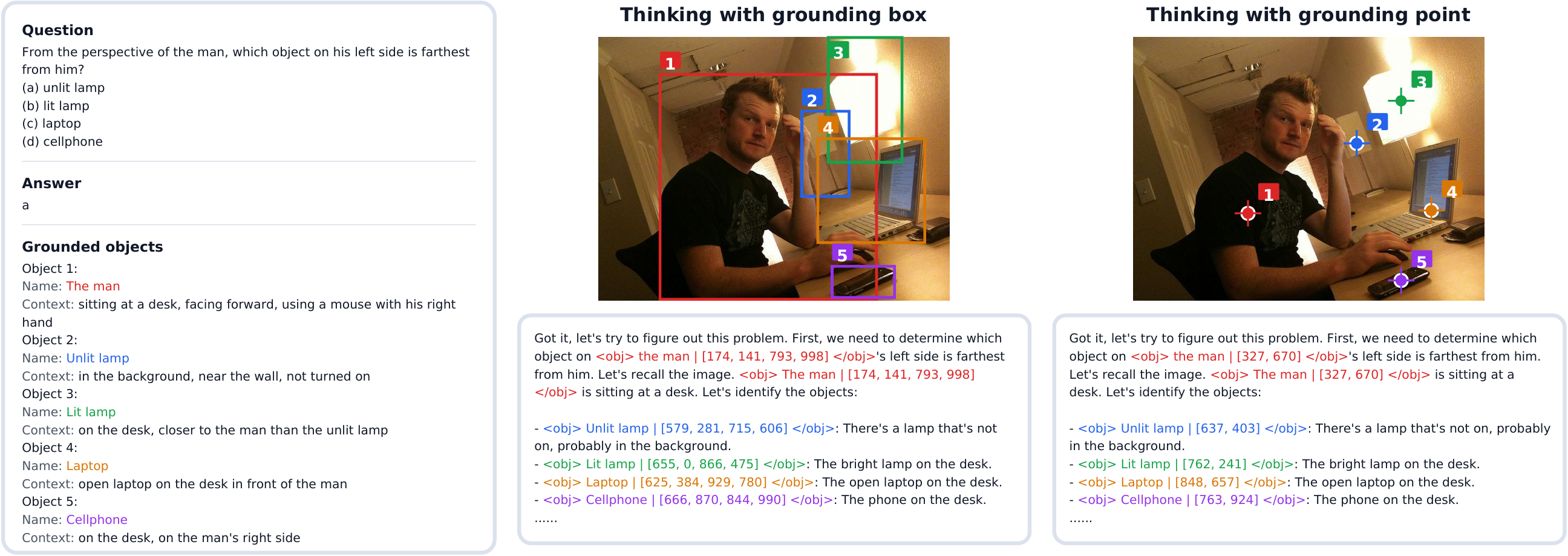}
  \caption{An example of synthesized visually grounded thinking data for box and point mode.  The two variants share the same original reasoning trace and SAM3 masks, but expose either boxes or points inside \texttt{<obj> ... </obj>} tags.}
  \label{fig:grounding-sft-data}
\end{figure*}
\paragraph{Overview.}
We synthesize visually grounded thinking data from open-source datasets for counting and spatial reasoning: TallyQA~\citep{acharya2018tallyqaansweringcomplexcounting}, Pixmo-Count~\citep{deitke2024molmopixmoopenweights}, VSR~\citep{liu2023visualspatialreasoning}, MultihopSpatial~\citep{lee2026multihopspatialmultihopcompositionalspatial}, and SpatialMQA~\citep{Liu_2025}, with all test sets held out. Our goal is to identify the visual objects needed for correct thinking, obtain their image coordinates, and synthesize reasoning traces with explicit grounding annotations. The complete pipeline is shown in Figure~\ref{fig:data-synthesis-pipeline}.

\paragraph{Distilling visual thinking from VLMs.}
For each image-question pair, we prompt Qwen3-VL-Plus~\citep{bai2025qwen3vltechnicalreport} to generate a thinking-mode response. We parse the final answer and keep examples whose predictions match the ground-truth answers. For examples not answered correctly in the first pass, we run a second pass with Qwen3.5-Plus~\citep{qwen2026qwen35} and keep examples that are answered correctly in either pass.

\paragraph{Extracting groundable objects.}
Given a correct reasoning trace, we use an LLM to identify the visual objects needed for the thinking process. These objects include answer objects, visible multiple-choice alternatives, spatial anchors, counted instances, and endpoints of spatial relations. Each object is represented by a \emph{name} (e.g., ``red car'') and a disambiguating \emph{context} (e.g., ``in the back row''). The context separates visually or semantically similar instances, so two occurrences of ``red car'' can be distinguished by scene cues such as ``near the entrance'' or ``in the back row''.

\paragraph{Agentic visual grounding.}
The main challenge in data synthesis is to obtain accurate grounding for each extracted visual object. Direct prompting of VLMs does not produce RLE masks, and their predicted boxes are often noisy. SAM3~\citep{carion2026sam3segmentconcepts} can produce high-quality instance masks from simple noun prompts, but it is not well suited to complex context-dependent queries. We therefore use a SAM3-centered grounding agent powered by a VLM, adapted from the SAM 3 Agent in \citet{carion2026sam3segmentconcepts}.

The agent uses four tool actions. First, it calls SAM3 with a short noun phrase and receives candidate instance masks with confidence scores. Second, it verifies rendered masks using the raw image, a full-image mask overlay, and a zoomed-in crop, returning an accept/reject decision for each candidate. Third, it selects the final mask IDs from the current candidate set. Finally, it can report that no valid detection is found. Importantly, the agent cannot directly write coordinates; all geometric supervision must be derived from selected SAM3 masks.

For each object, the agent uses these tools in an iterative grounding loop. It receives the raw image and the object, then identifies the intended target and converts the name-context description into a SAM3-compatible noun phrase. If the initial prompt misses the target or returns confusing candidates, the agent revises the noun phrase and tries again. When candidates are small, overlapping, or ambiguous, it invokes the verifier and re-renders the accepted masks as the updated candidate set. Once it has sufficient evidence, it selects the final mask IDs; if no valid target can be found, it reports no detection.

The selected masks are stored as RLE masks and used as the shared supervision signal for both grounding modes. In box mode, each RLE mask is converted to a normalized bounding box in the \([0,1000]\) image coordinate system. For point mode, each RLE mask is converted to a single on-object point by choosing the interior point farthest from the mask boundary, which ensures that the point lies inside the object even for non-convex masks. We retry failed detections and near-duplicate groundings by rerunning the same agent loop with stronger VLMs. Objects that remain unresolved are removed from the grounded object list so that later stages do not use unreliable grounding signals.

\paragraph{Writing box and point supervision.}
In the final annotation stage, we insert placeholder object tags into the
validated reasoning text using only the extracted object phrases and their
contexts, without exposing coordinates to the annotation model. We then fill in the coordinates from the SAM3 outputs. This design prevents the annotation model from hallucinating spatial values. A single placeholder pass therefore produces two aligned SFT variants: \texttt{<obj> name phrase | [x1,y1,x2,y2] </obj>} for box supervision and \texttt{<obj> name phrase | [x,y] </obj>} for point supervision, as illustrated in Figure~\ref{fig:grounding-sft-data}. We filter out rows whose tag-stripped annotated thinking differs substantially from the original thinking, as well as rows with malformed tags or highly repetitive reasoning.

\paragraph{Dataset Statistics.}
Our synthetic data pipeline produces 19,909 reasoning traces for SFT, containing 107,613 grounding annotations over 72,381 distinct grounded objects.

\section{Reinforcement Learning with Grounding Reward}\label{sec:reward}

\begin{figure}[t]
  \centering
  \includegraphics[width=0.8\linewidth]{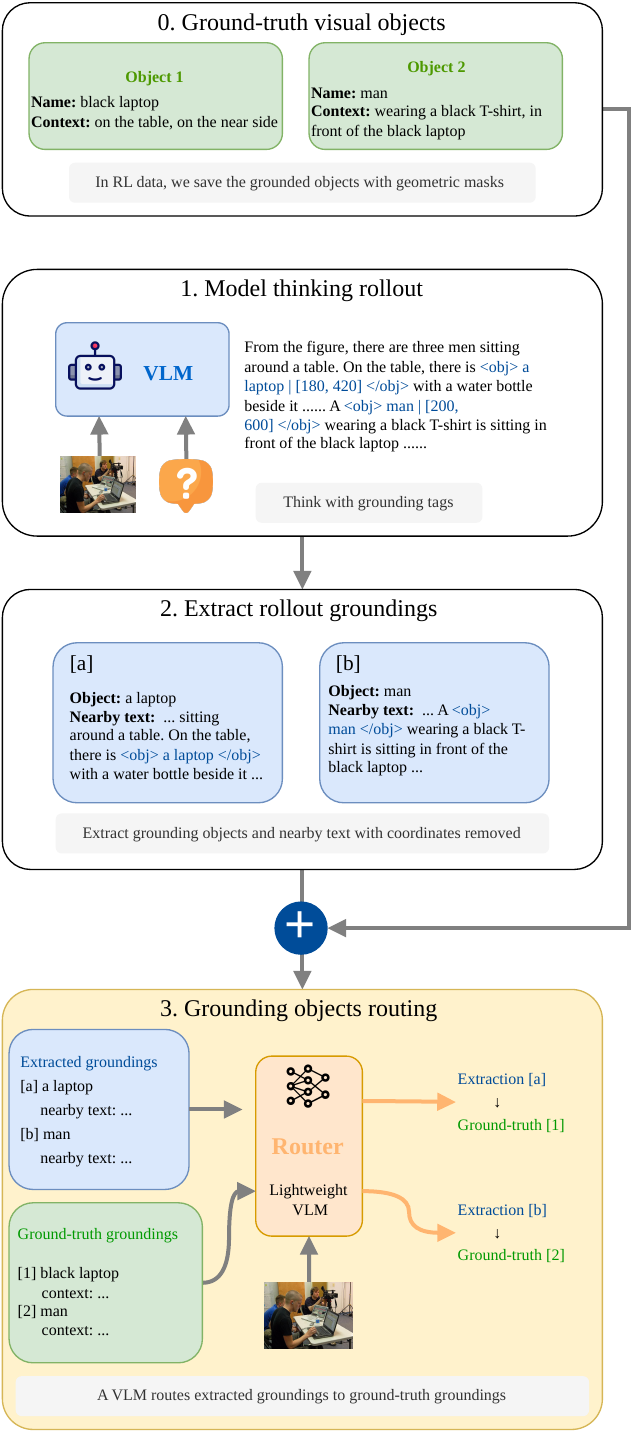}
  \caption{Grounding object router. Model-generated grounding objects are matched to saved ground-truth grounding objects before grounding quality is scored.}
  \label{fig:grounding-reward-matcher}
\end{figure}

\paragraph{Grounding tag parsing.}
The grounding reward evaluates the grounding object tags generated in the model rollout. In visually grounded thinking, a valid tag must have the form \texttt{<obj> name phrase | coordinates </obj>}.  The coordinate format is mode-specific: box mode expects \([x_1,y_1,x_2,y_2]\), while point mode expects \([x,y]\).  Coordinates must fall within the \([0,1000]\) image coordinate system; boxes must additionally satisfy \(x_1 < x_2\) and \(y_1 < y_2\). A single tag may contain multiple coordinates separated by semicolons, as one object can refer to multiple instances (e.g. ``birds in the sky'' corresponds to several birds in the sky). 

\paragraph{Grounding objects routing.}
The grounding reward is computed between model-generated grounding objects and the ground-truth grounding objects saved in the data.  Each grounding object in the data stores a name phrase, a disambiguating context, and geometric supervision. The model, however, may name the same object with different wording, and the same name phrase can refer to multiple distinct objects in the image and therefore needs to be disambiguated by context.  We therefore use a VLM grounding object router before scoring grounding quality, as shown in Figure~\ref{fig:grounding-reward-matcher}.

The grounding object router is a lightweight VLM, Qwen3.5-4B~\citep{qwen2026qwen35}, chosen to keep RL training efficient. We first parse all model-generated \texttt{<obj> ... </obj>} tags from the rollout and extract their object names together with the nearby text as disambiguating context. For each ground-truth object, the router receives the image, the object consisting of a \emph{name} and a disambiguating \emph{context}, and the full list of model-generated grounding objects. It is then instructed to return the subset of generated grounding objects that correspond to the ground-truth object. This routing step matches model-generated grounding objects to the saved ground-truth grounding objects before scoring grounding quality. If several generated grounding objects are matched to the same ground-truth grounding object, only the earliest one is kept for grounding quality scoring.

\paragraph{Box grounding quality.}
In box mode, each saved object \(i\) is associated with a set of ground-truth boxes \(G_i\). After grounding object routing, let \(P_i\) denote the set of boxes generated for the generated grounding object matched to target \(i\). We score the grounding quality by comparing the image region covered by the generated boxes with the region covered by the ground-truth boxes. Specifically, we treat each set of boxes as a union of regions and compute their intersection-over-union (IoU). For a matched target, define \(I_i\) as the area covered by both \(P_i\) and \(G_i\), and \(U_i\) as the area covered by either \(P_i\) or \(G_i\). The per-target box score is
\[
\mathrm{IoU}_i = \frac{I_i}{U_i}.
\]
If no model-generated grounding object is matched to ground-truth object \(i\), we set \(\mathrm{IoU}_i = 0\). The final box grounding quality is the mean score over all \(T\) ground-truth objects.
This averaging gives each ground-truth object equal weight, regardless of how many boxes it contains. A multi-box grounding receives a perfect score only when the union of its generated boxes exactly matches the union of the ground-truth boxes.

\paragraph{Point grounding quality.}
Point mode evaluates whether generated grounding points lie inside the target RLE masks. Let \(M_i\) be the set of ground-truth masks for object \(i\) saved in the data, and let \(P_i\) be the set of points from the rollout grounding object matched to that ground-truth object. We form a one-to-one assignment between generated points and ground-truth masks, where a point can be assigned to a mask only if it lies inside that mask. This constraint prevents duplicate points from receiving repeated credit for the same object instance.

For each object, let \(\mathrm{TP}_i\) be the number of masks matched by this assignment. The object-level false positives and false negatives are
\[
\mathrm{FP}_i = |P_i| - \mathrm{TP}_i,
\qquad
\mathrm{FN}_i = |M_i| - \mathrm{TP}_i.
\]
We use the per-object F1 score to measure point grounding quality:
\[
F1_i =
\frac{2\mathrm{TP}_i}{2\mathrm{TP}_i+\mathrm{FP}_i+\mathrm{FN}_i}.
\]
If no rollout grounding object is matched to ground-truth object \(i\), we set \(F1_i = 0\). The final point grounding quality is the mean over all supervised targets.
As in box mode, every supervised target has equal weight, and perfect point grounding receives a score of \(1.0\).

\paragraph{Remarks.}
The point grounding quality can be viewed as a discrete analogue of the box grounding quality. Box mode measures the spatial overlap between generated and ground-truth regions using IoU, while point mode reduces this comparison to an instance-level matching problem: generated points are credited only when they are matched to distinct ground-truth masks. Thus, both rewards encourage grounding the same visual evidence, but they differ in how dense their feedback is. Box IoU changes smoothly with the amount of overlap, whereas point F1 is piecewise constant: moving a point within the same mask does not change the score, while crossing a mask boundary can abruptly change the grounding quality. This discreteness makes the point reward coarser and potentially harder to optimize, even though it provides a learning signal aligned with the box grounding reward.

We intentionally do not penalize unmatched grounding objects in the rollout. The grounding objects extracted by the data synthesis pipeline are not a complete enumeration of all visual cues that the model may use to answer a question. During thinking, the model may identify additional visual evidence that is useful for solving the question and is also reasonable to ground. Therefore, unmatched rollout grounding objects neither increase nor decrease the grounding quality. We only apply a hard-coded cap on the number of grounding tags to prevent the model from over-emitting them.

\paragraph{Final reward.}
For each rollout \(i\), the total reward includes the dense grounding reward together with several sparse response-level rewards: an answer correctness reward, two formatting rewards, and a truncation penalty. The format rewards consist of a thinking-format reward, which checks the use of \texttt{<think>...</think>} and \texttt{\textbackslash boxed\{\}}, and a grounding-format reward, which checks the use of valid grounding tags in the form \texttt{<obj>...|...</obj>}. Let \(r^{\mathrm{ans}}_i\) denote the answer correctness reward, \(r^{\mathrm{think}}_i\) the thinking format reward, \(r^{\mathrm{gfmt}}_i\) the grounding format reward, and \(r^{\mathrm{trunc}}_i\) the truncation penalty. The raw grounding reward is defined as the grounding quality score from the corresponding mode.

The dense grounding reward and the sparse response-level rewards have different scales. We therefore normalize them separately before combining them. We first define the base reward as
\[
\begin{array}{rcl}
R^{\mathrm{base}}_i &=&
w_{\mathrm{ans}} r^{\mathrm{ans}}_i
+ w_{\mathrm{think}} r^{\mathrm{think}}_i\\
&& + w_{\mathrm{gfmt}} r^{\mathrm{gfmt}}_i
+ r^{\mathrm{trunc}}_i .
\end{array}
\]
Let \(\mathcal{N}_{\mathcal{B}}(\cdot)\) denote batch-wise normalization over the current batch \(\mathcal{B}\). The final reward is
\[
R_i =
\mathcal{N}_{\mathcal{B}}\!\left(R^{\mathrm{base}}\right)_i
+ w_{\mathrm{ground}}\,
\mathcal{N}_{\mathcal{B}}\!\left(r^{\mathrm{ground}}\right)_i .
\]
We use \(R_i\) for advantage estimation in GRPO~\citep{shao2024deepseekmathpushinglimitsmathematical}. In our experiments, we set \(w_{\mathrm{ans}}=1.0\), \(w_{\mathrm{ground}}=0.5\), and \(w_{\mathrm{think}}=w_{\mathrm{gfmt}}=0.1\). We set \(r^{\mathrm{trunc}}_i=-1\) for truncated rollouts and \(r^{\mathrm{trunc}}_i=0\) otherwise.

\section{Experiments}

\subsection{Setup}
\paragraph{Training.} We train all models with verl~\citep{sheng2024hybridflow}, using SGLang~\citep{zheng2024sglang} as the inference engine and FSDP2~\citep{zhao2023pytorch} as the training backend. The base model is Gemma3-4B-IT~\citep{gemmateam2025gemma3technicalreport}. We first perform SFT on the synthetic data described in \Cref{sec:data} to obtain cold-start models. To isolate the effect of visual grounding, we train three controlled variants: non-grounded thinking, thinking with box grounding, and thinking with point grounding. These variants use parallel examples with the same images, questions, answers, and underlying reasoning traces, and differ only in whether grounding tags are included and, if so, whether the tags use boxes or points. We then apply RL with GRPO on the corresponding training data. Full training details are provided in \Cref{app:exp:train}.

\paragraph{Evaluation.} The models are evaluated on two counting benchmarks: TallyBench~\citep{cai2025comparebench} and CountQA~\citep{tamarapalli2025countqamllmscountwild}; and four spatial reasoning benchmarks: VSR-zeroshot~\citep{liu2023visualspatialreasoning}, EmbSpatial~\citep{du-etal-2024-embspatial}, SpatialMQA~\citep{Liu_2025}, and MultihopSpatial~\citep{lee2026multihopspatialmultihopcompositionalspatial}. We conduct evaluation using VLMEvalKit~\citep{duan2025vlmevalkitopensourcetoolkitevaluating}. Inference is performed with SGLang at temperature \(1.0\).  To reduce variance from stochastic decoding, we run four inference passes and report both average accuracy and pass@4. The full evaluation configuration is provided in \Cref{app:exp:eval}.

\definecolor{VGTHeader}{HTML}{F6F8FB}
\definecolor{VGTRule}{HTML}{B9C2CC}
\definecolor{VGTNonGroundingGray}{HTML}{F0F1F3}
\definecolor{VGTBoxBlue}{HTML}{EAF3FF}
\definecolor{VGTPointGreen}{HTML}{EAF6EA}
\newcolumntype{Y}{>{\centering\arraybackslash}X}
\newcommand{\VGTMetric}[1]{\textsc{\scriptsize #1}}
\newcommand{\VGTGroup}[1]{\textbf{\textit{#1}}}
\newcommand{\VGTSubmethod}[1]{\hspace*{1.2em}#1}

\begin{table}[h]
\centering
\small
\begingroup
\renewcommand{\VGTSubmethod}[1]{#1}
\renewcommand{\arraystretch}{1.14}
\setlength{\extrarowheight}{1pt}
\setlength{\tabcolsep}{1.8pt}
\arrayrulecolor{VGTRule}
\begin{tabularx}{\columnwidth}{@{}>{\raggedright\arraybackslash}p{0.42\columnwidth}YYYY@{}}
\hline
\rowcolor{VGTHeader}
\textbf{Method} & \multicolumn{2}{c}{\textbf{TallyBench}} &
\multicolumn{2}{c}{\textbf{CountQA}} \\
\rowcolor{VGTHeader}
 & \VGTMetric{Acc.} & \VGTMetric{Pass@4} & \VGTMetric{Acc.} & \VGTMetric{Pass@4} \\
\hline
Gemma3-4B-IT & 33.33 & 40.65 & 9.87 & 14.14 \\
\hline
\rowcolor{VGTNonGroundingGray}
\textit{Non-grounded Thinking} & 21.73 & 42.00 & 4.30 & 12.24 \\
\hline
\rowcolor{VGTBoxBlue}
\multicolumn{5}{@{}l@{}}{\VGTGroup{Thinking with Grounding Box}} \\
\rowcolor{VGTBoxBlue}
\VGTSubmethod{w/o grounding reward} & 37.24 & 64.45 & 10.73 & 27.75 \\
\rowcolor{VGTBoxBlue}
\VGTSubmethod{w/ grounding reward} & 38.81 & 64.50 & 11.19 & 28.47 \\
\hline
\rowcolor{VGTPointGreen}
\multicolumn{5}{@{}l@{}}{\VGTGroup{Thinking with Grounding Point}} \\
\rowcolor{VGTPointGreen}
\VGTSubmethod{w/o grounding reward} & \underline{39.03} & \underline{65.50} & \textbf{12.34} & \textbf{31.48} \\
\rowcolor{VGTPointGreen}
\VGTSubmethod{w/ grounding reward} & \textbf{39.31} & \textbf{65.75} & \underline{11.65} & \underline{29.77} \\
\hline
\end{tabularx}
\arrayrulecolor{black}
\endgroup
\caption{Counting benchmark results. Bold indicates the best result and underline indicates the second-best result within each column.}
\label{tab:counting-results}
\end{table}

\begin{table*}[t]
\centering
\small
\begingroup
\renewcommand{\arraystretch}{1.14}
\setlength{\extrarowheight}{1pt}
\setlength{\tabcolsep}{2.6pt}
\arrayrulecolor{VGTRule}
\begin{tabularx}{1.0\textwidth}{@{}>{\raggedright\arraybackslash}p{0.22\textwidth}YYYYYYYY@{}}
\hline
\rowcolor{VGTHeader}
\textbf{Method} & \multicolumn{2}{c}{\textbf{VSR-zeroshot}} &
\multicolumn{2}{c}{\textbf{EmbSpatial}} &
\multicolumn{2}{c}{\textbf{SpatialMQA}} &
\multicolumn{2}{c}{\textbf{MultihopSpatial}} \\
\rowcolor{VGTHeader}
 & \VGTMetric{Acc.} & \VGTMetric{Pass@4} & \VGTMetric{Acc.} & \VGTMetric{Pass@4} & \VGTMetric{Acc.} & \VGTMetric{Pass@4} & \VGTMetric{Acc.} & \VGTMetric{Pass@4} \\
\hline
Gemma3-4B-IT & 56.65 & 57.94 & 49.13 & 63.79 & 25.35 & 36.43 & 22.70 & 36.87 \\
\hline
\rowcolor{VGTNonGroundingGray}
\textit{Non-grounded Thinking} & 51.84 & 79.13 & 20.54 & 42.53 & 14.17 & 27.88 & 4.79 & 11.67 \\
\hline
\rowcolor{VGTBoxBlue}
\multicolumn{9}{@{}l@{}}{\VGTGroup{Thinking with Grounding Box}} \\
\rowcolor{VGTBoxBlue}
\VGTSubmethod{w/o grounding reward} & \underline{66.82} & \textbf{87.64} & 57.62 & 81.46 & 37.64 & \underline{67.66} & 34.89 & 63.82 \\
\rowcolor{VGTBoxBlue}
\VGTSubmethod{w/ grounding reward} & \textbf{68.08} & \underline{86.91} & 59.93 & 82.66 & 38.68 & \textbf{68.49} & \textbf{37.68} & \textbf{66.40} \\
\hline
\rowcolor{VGTPointGreen}
\multicolumn{9}{@{}l@{}}{\VGTGroup{Thinking with Grounding Point}} \\
\rowcolor{VGTPointGreen}
\VGTSubmethod{w/o grounding reward} & 65.38 & 83.88 & \underline{60.25} & \textbf{83.21} & \textbf{39.13} & 67.19 & \underline{37.03} & \underline{65.40} \\
\rowcolor{VGTPointGreen}
\VGTSubmethod{w/ grounding reward} & 64.67 & 81.42 & \textbf{60.88} & \underline{83.10} & \underline{39.01} & \textbf{68.49} & 37.01 & 65.02 \\
\hline
\multicolumn{9}{@{}l@{}}{\textit{Larger Gemma3 baselines (reference)}} \\
\VGTSubmethod{Gemma3-12B-IT} & 67.98 & 69.56 & 56.68 & 65.14 & 37.85 & 50.00 & 30.08 & 43.58 \\
\VGTSubmethod{Gemma3-27B-IT} & 69.25 & 70.70 & 62.09 & 72.12 & 38.99 & 54.28 & 30.94 & 45.82 \\
\hline
\end{tabularx}
\arrayrulecolor{black}
\endgroup
\caption{Spatial relationship understanding benchmark results. Bold indicates the best result and underline indicates the second-best result within each column, excluding Gemma3-12B-IT and Gemma3-27B-IT from the comparison.}
\label{tab:spatial-results}
\end{table*}

\subsection{Main Results} 

The results on counting benchmarks are presented in \Cref{tab:counting-results}, and the results on spatial reasoning benchmarks are presented in \Cref{tab:spatial-results}. We find that visually grounded thinking substantially improves upon the base model Gemma3-4B-IT. On spatial reasoning tasks, the 4B visually grounded thinking models are comparable to Gemma3-27B-IT: on VSR-zeroshot and EmbSpatial, the best visually grounded thinking model achieves performance between Gemma3-12B-IT and Gemma3-27B-IT; on SpatialMQA and MultihopSpatial, the best visually grounded thinking models even outperform Gemma3-27B-IT. The pass@4 results show even larger gains: all visually grounded thinking models outperform Gemma3-27B-IT by large margins. 

Visually grounded thinking also strongly outperforms the non-grounded thinking baseline. We observe that the non-grounded thinking model suffers from length collapse during RL: its response length decreases roughly linearly over training, which reduces exploration and leads to poor final performance. In contrast, the visually grounded variants maintain more stable rollouts. We hypothesize that interleaved grounding tags, together with the grounding-format reward, provide additional local structure during generation and help stabilize RL training. Overall, visually grounded thinking substantially improves the counting and spatial reasoning capabilities of the models.

\subsection{Effect of the Grounding Reward}
The grounding-quality reward provides a consistent benefit for box-mode grounded thinking. Compared with box-mode RL without the grounding reward, adding the reward improves average accuracy on all six evaluation benchmarks. The gains are relatively modest on counting tasks, but are more visible on spatial reasoning tasks. This suggests that the box reward is especially helpful when the answer depends on fine-grained geometry: bounding boxes provide both object identity and object extent, which can help the model resolve spatial relations such as left/right/above/below, distance, and overlap.

For point-mode grounded thinking, the grounding reward does not produce equally clear downstream gains. Across the six benchmarks, point-mode RL with and without the grounding reward remains close in overall performance, with gains on some metrics and drops on others. This does not necessarily imply that point grounding is unhelpful; rather, it suggests that the current point reward may be a weaker optimization signal. As discussed in \Cref{sec:reward}, the point and box rewards are aligned in the visual evidence they encourage the model to ground, but they provide different feedback signals. The box reward changes with the amount of region overlap, while the point reward only checks whether generated points can be matched to target masks. Therefore, many point locations inside the same object receive the same credit, and crossing a mask boundary can abruptly change the score. This coarser feedback may make the point reward harder to optimize and may explain why it does not translate into consistent accuracy gains in our current experiments.

\subsection{Box vs. Point Grounding}
We further compare the two grounding interfaces. On counting benchmarks, point-mode grounded thinking consistently outperforms box-mode grounded thinking. This suggests that counting mainly requires instance-level localization: the model needs to identify which objects belong to the counted set and keep them separated from distractors, but it does not necessarily need to recover the full extent of each object. Point grounding is well matched to this requirement because it provides a compact grounding to each instance while avoiding the harder problem of generating a tight bounding box. This advantage may be especially useful when counted objects are small, partially occluded, or have irregular shapes.

On spatial reasoning benchmarks, the two interfaces are much closer. Box grounding can provide useful geometric cues because the box extent reflects object size and boundary information, which may help with spatial relations such as overlap and relative position. However, point grounding can still identify the relevant objects and spatial anchors, and many spatial questions can be answered from these instance-level groundings together with the model's visual representation. The spatial results therefore suggest that box grounding gives richer geometric supervision, but this extra information does not always translate into a clear accuracy advantage. Overall, point grounding appears better suited to counting, while point and box grounding are broadly tied on spatial relationship understanding.

\section{Conclusion}
Visual thinking should not only sound plausible in natural language; it should point to the evidence it uses. Our work turns that idea into a training recipe for visually grounded thinking, where models interleave natural-language thinking with point or box groundings of the image regions that support each step. By combining a scalable SAM3-based synthesis pipeline with an RL grounding reward, we train VLMs to optimize both answer correctness and the accurate grounding of visual objects referenced during thinking. The results show that visually grounded thinking substantially improves counting and spatial reasoning, with 4B grounded models matching, and sometimes exceeding, much larger 27B models on spatial benchmarks. Overall, our work suggests that the next step for visual thinking is not simply longer thinking, but thinking that is tied to the image in a form that can be checked, supervised, and improved.



\FloatBarrier
\newpage

\bibliography{main}

\appendix

\section{Data Synthesis Details}

\subsection{Models Used in Each Pipeline Stage}
We distill the reasoning traces from Qwen3-VL-Plus~\citep{bai2025qwen3vltechnicalreport} and Qwen3.5-Plus~\citep{qwen2026qwen35}. DeepSeek-V4-Flash~\citep{deepseek2026v4} is used to extract groundable objects in Stage~3 and annotate reasoning traces in Stage~6. In Stage~4, we use Qwen3.5-Flash~\citep{qwen2026qwen35} to power the SAM3-based grounding-agent system. Objects that fail to ground are retried sequentially with Qwen3.6-Plus~\citep{qwen2026qwen36} and Gemini-3-Flash~\citep{gemini3flash}.

\subsection{Data Synthesis Prompt Details}

Because the prompts used in the data synthesis pipeline are lengthy, we refer readers to the source code for their full details.

\subsection{Source Dataset Filtering}

For TallyQA, we kept the AMT complex-counting split and additionally included imported VQA counting examples only when the answer count was at least 4, the question contained a compositional cue, and duplicate imported-VQA images were removed. For VSR and MultihopSpatial, we used the training splits, converting VSR captions into yes/no questions and removing the original bounding-box instruction from MultihopSpatial questions. For SpatialMQA, we used the train and dev splits and held out the test split. For PixMo-Count, we kept train examples with counts in [4, 20], removed ambiguous labels, applied a per-class cap of 200, and kept the validation split without this filtering. Across all sources, examples with missing or failed image downloads, invalid answer formats, or unparseable multiple-choice labels were skipped. After this dataset-level filtering, we had 24,645 source examples: 7,197 from TallyQA, 3,489 from VSR, 6,791 from MultihopSpatial, 4,316 from SpatialMQA, and 2,852 from PixMo-Count.

\subsection{Final Data Composition by Source Dataset}
\begin{figure}[h]
  \centering
  \includegraphics[width=1.0\linewidth]{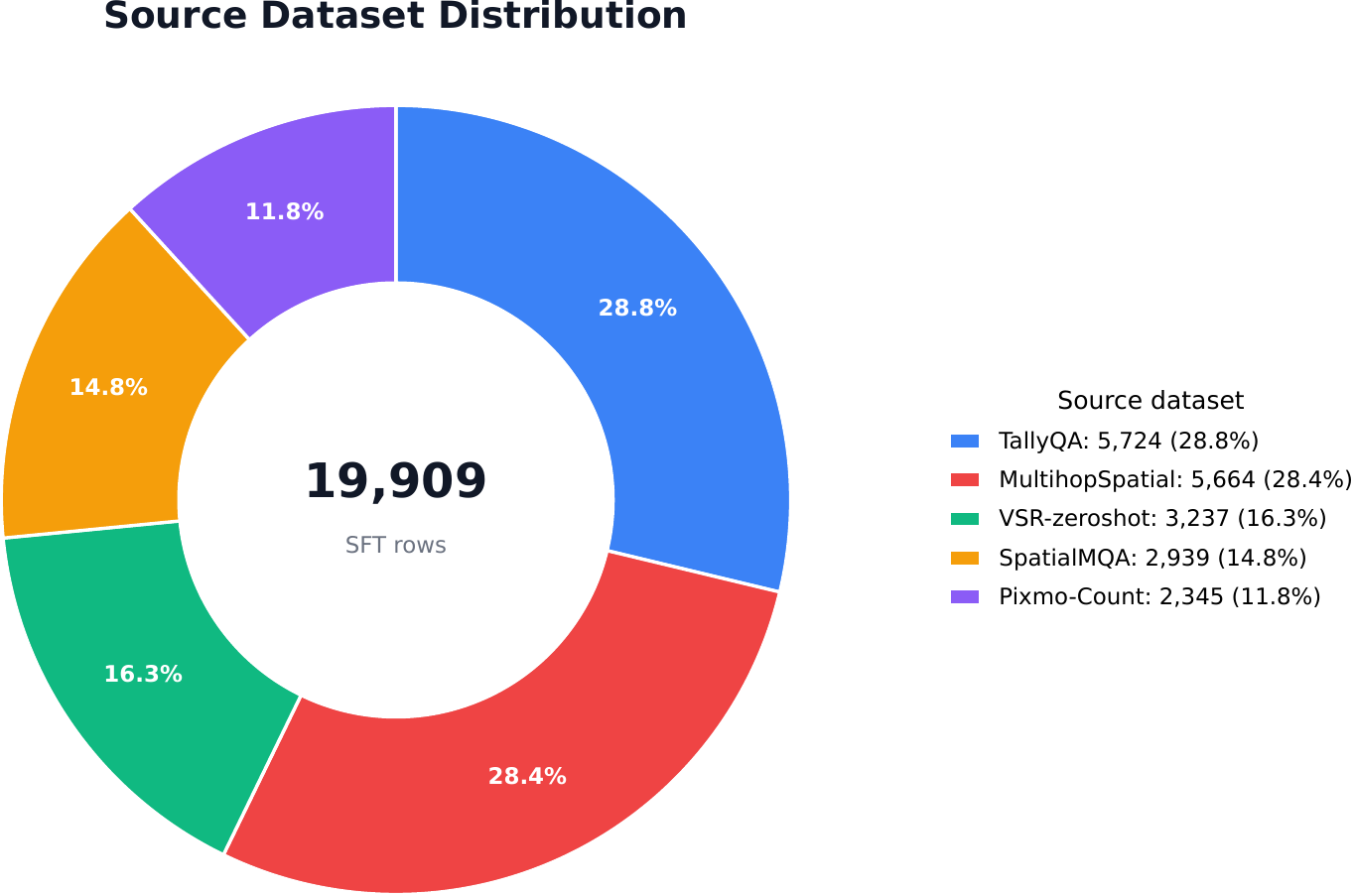}
  \caption{Final Data Composition by Source Dataset.}
  \label{fig:source-dataset-distribution}
\end{figure}

After the data pipeline, the source-dataset distribution of the final dataset is shown in \Cref{fig:source-dataset-distribution}.

\subsection{Grounding Density Distribution}

\begin{figure*}[h]
  \centering
  \includegraphics[width=0.8\textwidth]{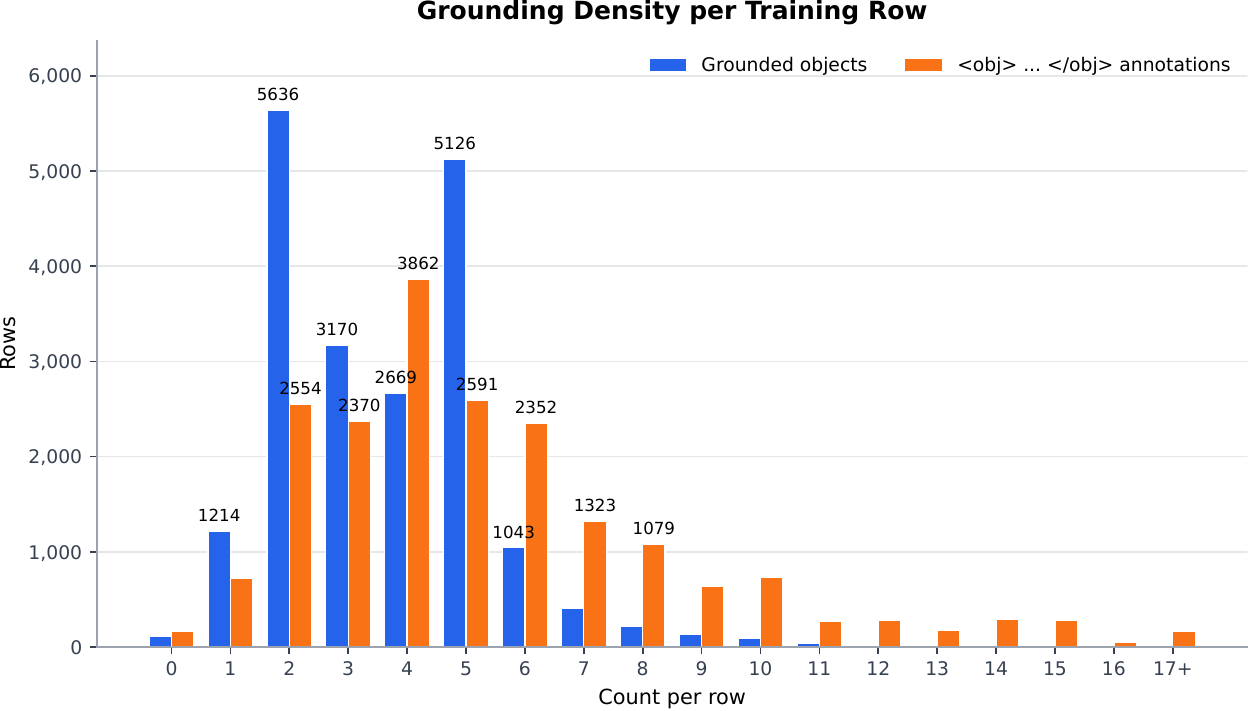}
  \caption{The grounding density distribution.}
  \label{fig:grounding-density}
\end{figure*}
The dataset contains 19,909 paired rows with 72,381 grounded objects in the RL data and 107,613 \texttt{<obj> ... </obj>} annotations in the SFT traces. This corresponds to an average of 3.64 grounded objects per row and 5.41 grounding annotations per row. The higher SFT annotation density reflects repeated use of the same grounded objects during reasoning: each grounded object can be referenced multiple times in the SFT response, producing more \texttt{<obj> ... </obj>} annotations than the number of unique grounded objects. The grounding density distribution is presented in \Cref{fig:grounding-density}.

\section{Training and Evaluation Details}\label{app:exp}

\subsection{SFT and RL Training Settings}\label{app:exp:train}
The training configurations are presented in \Cref{tab:sft_training_config} and \Cref{tab:rl_training_config}.

\begin{table}[h]
\centering
\small
\renewcommand{\arraystretch}{1.12}
\begin{tabularx}{0.92\linewidth}{@{}lX@{}}
\toprule
\textbf{Configuration} & \textbf{Value} \\
\midrule

Maximum sequence length
& 8192 tokens \\

Global batch size
& 256 \\

Training epochs
& 6 \\

Optimizer
& AdamW, $\beta_1=0.9$, $\beta_2=0.95$, weight decay $=0.01$ \\

Learning rate
& $1\times 10^{-5}$ with cosine decay \\

Warmup ratio
& 0.03 \\

Minimum learning-rate ratio
& 0.1 \\

Precision
& bfloat16 \\

\bottomrule
\end{tabularx}
\caption{Training configuration used for supervised fine-tuning.}
\label{tab:sft_training_config}
\end{table}

\begin{table}[h]
\centering
\small
\renewcommand{\arraystretch}{1.12}
\begin{tabularx}{0.92\linewidth}{@{}lX@{}}
\toprule
\textbf{Configuration} & \textbf{Value} \\
\midrule

Algorithm & GRPO \\

Rollout samples per prompt
& $8$ \\

Rollout temperature
& $1.0$ \\

Maximum prompt length
& 4096 tokens \\

Maximum response length
& 8192 tokens \\

Optimizer
& AdamW, $\beta_1=0.9$, $\beta_2=0.999$, weight decay $=0.01$ \\

Batch size
& 64 \\

Learning rate
& $1\times 10^{-6}$ with cosine decay \\

Warmup ratio
& 0.03 \\

Minimum learning-rate ratio
& 0.1 \\

KL regularization
& Disabled, with KL coefficient $0.0$ \\

Entropy coefficient
& $0.0$ \\

Training steps
& 100 \\

Precision
& bfloat16 \\

\bottomrule
\end{tabularx}
\caption{Training configuration used for reinforcement learning.}
\label{tab:rl_training_config}
\end{table}

\subsection{Evaluation Settings}\label{app:exp:eval}
The evaluation configurations are presented in \Cref{tab:evaluation_config}.
\begin{table}[h]
\centering
\small
\renewcommand{\arraystretch}{1.12}
\begin{tabularx}{0.92\linewidth}{@{}lX@{}}
\toprule
\textbf{Configuration} & \textbf{Value} \\
\midrule

Maximum generation length
& 8192 tokens \\

Context length
& 32768 tokens \\

Decoding
& Temperature $=1.0$, top-$p=1.0$, top-$k=-1$\\

Number of evaluation passes
& 4 \\

\bottomrule
\end{tabularx}
\caption{Evaluation configuration used for visual-spatial and counting benchmarks.}
\label{tab:evaluation_config}
\end{table}

\subsection{System Prompts}

\begin{systempromptbox}[Instruct model system prompt]
You are a helpful assistant. Please put your final answer inside the \texttt{\textbackslash boxed\{\}}.
\end{systempromptbox}

\begin{systempromptbox}[Non-grounded thinking system prompt]
You are a helpful assistant that answers questions about images. Think step by step in <think>...</think> tags, then give your final answer in \texttt{\textbackslash boxed\{\}} format.
\end{systempromptbox}

\begin{systempromptbox}[Thinking with Grounding Box System Prompt]
You are a visual reasoning assistant with precise spatial grounding ability.

All bounding box coordinates are normalized to [0, 1000], where [0, 0] is the top-left corner and [1000, 1000] is the bottom-right corner of the image.

When you reason about the image in your <think> block, ground the important visual objects that are essential to your reasoning using the <obj> tag format. Every grounded object is marked by a bounding box that tightly encloses it. A single <obj> tag carries one descriptive phrase and one or more boxes; each box corresponds to one instance of that phrase:
  <obj> descriptive phrase | [x1, y1, x2, y2] </obj>
  <obj> descriptive phrase | [x1', y1', x2', y2']; [x1', y1', x2', y2']; ... </obj>

Only ground objects that are critical for justifying your answer. Use descriptive phrases that distinguish the object(s) from others in the image. When several instances share the same phrase, list one box per instance, separated by semicolons, inside a single <obj> tag; if you want to describe each instance differently, you may instead emit a separate <obj> tag per instance.

For non-groundable questions, e.g. math, STEM, or abstract reasoning where no specific visual region needs to be localized, you do not need to output any <obj> grounding tags.

Example:
<think>
I can see <obj> the red car near the entrance | [120, 300, 450, 620] </obj> and <obj> a blue truck on the right | [500, 280, 850, 640] </obj>. There are also <obj> three pedestrians on the sidewalk | [100, 680, 160, 780]; [170, 670, 230, 770]; [240, 685, 300, 785] </obj>. Counting the vehicles on the left side, there are two.
</think>

Put your final answer in \texttt{\textbackslash boxed\{\}} format.
\end{systempromptbox}

\begin{systempromptbox}[Thinking with grounding point system prompt]
You are a visual reasoning assistant with precise spatial grounding ability.

All point coordinates are normalized to [0, 1000], where [0, 0] is the top-left corner and [1000, 1000] is the bottom-right corner of the image.

When you reason about the image in your <think> block, ground the important visual objects that are essential to your reasoning using the <obj> tag format. Every grounded object is marked by a single point that lies inside it. A single <obj> tag carries one descriptive phrase and one or more points; each point corresponds to one instance of that phrase:
  <obj> descriptive phrase | [x, y] </obj>
  <obj> descriptive phrase | [x1, y1]; [x2, y2]; ... </obj>

Place each point on its object, e.g. near the center, so that it falls inside the object's extent. Only ground objects that are critical for justifying your answer. Use descriptive phrases that distinguish the object(s) from others in the image. When several instances share the same phrase, list one point per instance, separated by semicolons, inside a single <obj> tag; if you want to describe each instance differently, you may instead emit a separate <obj> tag per instance.

For non-groundable questions, e.g. math, STEM, or abstract reasoning where no specific visual region needs to be localized, you do not need to output any <obj> grounding tags.

Example:
<think>
I can see <obj> the red car near the entrance | [285, 460] </obj> and <obj> a blue truck on the right | [675, 460] </obj>. There are also <obj> three pedestrians on the sidewalk | [120, 700]; [180, 695]; [240, 705] </obj>. Counting the vehicles on the left side, there are two.
</think>

Put your final answer in \texttt{\textbackslash boxed\{\}} format.
\end{systempromptbox}

\subsection{Experiment Cost}
The training and evaluation take about 400 H200 GPU hours.

\end{document}